\pdfoutput=1

\documentclass[11pt]{article}

\usepackage{acl}

\usepackage{times}
\usepackage{latexsym}

\usepackage[T1]{fontenc}

\usepackage[utf8]{inputenc}

\usepackage{microtype}

\usepackage{enumitem}
\usepackage{url}
\usepackage{arydshln}
\usepackage{graphicx}
\usepackage{booktabs}
\usepackage{pifont}
\newcommand{\cmark}{\ding{51}}%
\newcommand{\xmark}{\ding{55}}%

\usepackage{algorithm}
\usepackage[noend]{algpseudocode}
\usepackage{xcolor, soul}
\usepackage{multirow, makecell}

\definecolor{Background}{RGB}{217,245,203}
\sethlcolor{Background}

\newif\ifcomments
\commentstrue 
\ifcomments
    \newcommand\tw[1]{\textcolor{red}{[TW: #1]}}
    \newcommand\jb[1]{\textcolor{blue}{[JB: #1]}}
    \newcommand\dd[1]{\textcolor{magenta}{[DD: #1]}}

\else
    \providecommand{\tw}[1]{}
    \providecommand{\jb}[1]{}
    \providecommand{\dd}[1]{}
\fi

\newcommand\breakdata{\textsc{Break}}
\newcommand\academic{\textsc{Academic}}

\newcommand\geo{\textsc{Geo880}}
\newcommand\imdb{\textsc{IMDB}}
\newcommand\yelp{\textsc{Yelp}}
\newcommand\spider{\textsc{Spider}}

\newcommand\tfive{T5-large}

\newcommand\tfivegold{T5-SQL-G}
\newcommand\tfivegoldpartial{T5-SQL-G\textsubscript{part}}
\newcommand\tfiveqdmr{T5-QDMR-G}

\newcommand\tfivepred{T5-QDMR-P}

\title{Weakly Supervised Text-to-SQL Parsing through Question Decomposition}

\author{\makecell{Tomer Wolfson$^{1,2}$ ~~~~~ Daniel Deutch$^{1}$ ~~~~~ Jonathan Berant$^{1}$ } \\ 
$^{1}$Blavatnik School of Computer Science, Tel Aviv University\hspace{5mm}
$^{2}$Allen Institute for AI \\ 
\texttt{\small\makecell{tomerwol@mail.tau.ac.il, danielde@post.tau.ac.il ,joberant@cs.tau.ac.il}}}

\begin{document}
\maketitle

\begin{abstract}

Text-to-SQL parsers are crucial in enabling non-experts to effortlessly query relational data. Training such parsers, by contrast, generally requires expertise in annotating natural language (NL) utterances with corresponding SQL queries.
In this work, we propose a \textit{weak supervision} approach for training text-to-SQL parsers. We take advantage of the recently proposed question meaning representation called QDMR, an intermediate between NL and formal query languages.
Given questions, their QDMR structures (annotated by non-experts or automatically predicted), and the answers, we are able to automatically synthesize SQL queries that are used to train text-to-SQL models. We test our approach by experimenting on five benchmark datasets. Our results show that the weakly supervised models perform competitively with those trained on annotated NL-SQL data.
Overall, we effectively train text-to-SQL parsers, while using zero SQL annotations.
\end{abstract}

\section{Introduction}
\label{sec:introduction}

The development of natural language interfaces to databases has been extensively studied in recent years \cite{Affolter2019ACS, Kim2020NaturalLT, Thorne2021FromNL}. 
The current standard is Machine Learning (ML) models which map utterances in natural language (NL) to executable SQL queries \cite{Wang2020RATSQLRS, rubin-berant-2021-smbop}.
These models rely on supervised training examples of NL questions labeled with their corresponding SQL queries. Labeling copious amounts of data is cost-prohibitive as it requires experts that are familiar both with SQL and with the underlying database structure \cite{Yu2018SpiderAL}.
Furthermore, it is often difficult to re-use existing training data in one domain in order to generalize to new ones \cite{Suhr2020ExploringUG}.
Adapting the model to a new domain requires new NL-SQL training examples, which results in yet another costly round of annotation. 


\begin{figure*}[t]\setlength{\belowcaptionskip}{-8pt}
  \centering
  \includegraphics[trim={0cm 9.2cm 1.25cm 0cm}, clip, width=\textwidth]{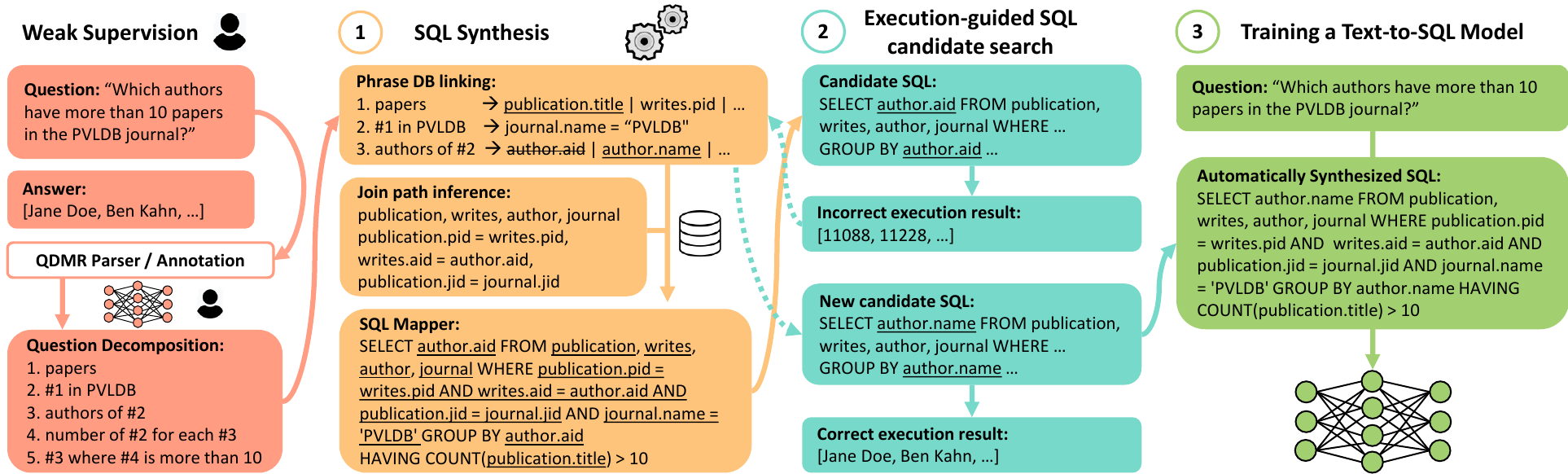}
  \vspace*{-0.7cm}
  \caption{Our pipeline for training a Text-to-SQL model on data synthesized using weak supervision.}
  \label{fig:architecture}
\end{figure*}

In this paper we propose a \emph{weak supervision} approach for training text-to-SQL parsers. We avoid the use of manually labeled NL-SQL examples and rely instead on data provided by non-expert users. Fig.~\ref{fig:architecture} presents a high-level view of our approach. The input (left corner, in red) is used to automatically synthesize SQL queries (step 3, in green) which, in turn, are used to train a text-to-SQL model. The supervision signal consists of the question's answer and uniquely, a structured representation of the \emph{question decomposition}, called QDMR. The annotation of both these supervision sources can be effectively crowdsourced to non-experts \cite{berant2013freebase,pasupat2015compositional, wolfson-etal-2020-break}. 
In a nutshell, QDMR is a series of computational steps, expressed by semi-structured utterances, that together match the semantics of the original question. The bottom left corner of Fig.~\ref{fig:architecture} shows an example QDMR of the question \textit{``Which authors have more than 10 papers in the PVLDB journal?''}. The question is broken into five steps, where each step expresses a single logical operation (e.g., select papers, filter those in PVLDB) and may refer to previous steps.
As QDMR is derived entirely from its question, it is agnostic to the underlying form of knowledge representation and has been used for questions on images, text and databases \cite{subramanian-etal-2020-obtaining, Geva2021BreakPB, Saparina2021SPARQLingDQ}. 
In our work, we use QDMR as an intermediate representation for SQL synthesis. Namely, we implement an automatic procedure that given an input QDMR, maps it to SQL. 
The QDMR can either be manually annotated or effectively predicted by a trained model, as shown in our experiments.

We continue to describe the main components of our system, using the aforementioned supervision (Fig.~\ref{fig:architecture}). 
The \emph{SQL Synthesis} component (step 1) attempts to convert the input QDMR into a corresponding SQL query. To this end, \emph{Phrase DB linking} matches phrases in the QDMR with relevant columns and values in the database. Next, SQL \textit{join paths} are automatically inferred given the database schema structure. Last, the QDMR, DB-linked columns and inferred join paths are converted to SQL by the \emph{SQL Mapper}. In step 2, we rely on question-answer supervision to filter out incorrect candidate SQL. Thus, our \emph{Execution-guided SQL Search} returns the first candidate query which executes to the correct answer. 

Given our synthesis procedure, we evaluate its ability to produce accurate SQL, using weak supervision.
To this end, we run our synthesis on 9,313 examples of questions, answers and QDMRs from five standard text-to-SQL benchmarks \cite{zelle96geoquery,  Li2014NaLIRAI, Yaghmazadeh2017SQLizerQS, Yu2018SpiderAL}.
Overall, our solution successfully synthesizes SQL queries for 77.8\% of examples, thereby demonstrating its applicability to a broad range of target databases.

Next, we show our synthesized queries to be an effective alternative to training on expert annotated SQL.
We compare a text-to-SQL model, trained on the queries synthesized from questions, answers and QDMRs, to one trained using gold SQL. As our model of choice we use \tfive{}, which is widely used for sequence-to-sequence modeling tasks \cite{Raffel2020ExploringTL}. Following past work \cite{Shaw2021CompositionalGA, Herzig2021UnlockingCG}, we fine-tune T5 to map text to SQL. 
We experiment with the \spider{} and \geo{} datasets \cite{Yu2018SpiderAL, zelle96geoquery} and compare model performance based on the training supervision.
When training on manually \emph{annotated QDMRs}, the weakly supervised models achieve 91\% to 97\% of the accuracy of models trained on gold SQL.
We further extend our approach to use automatically \emph{predicted QDMRs}, requiring zero annotation of in-domain QDMRs.
Notably, when training on predicted QDMRs models still reach 86\% to 93\% of the fully supervised versions accuracy.   
In addition, we evaluate cross-database generalization of models trained on \spider{} \cite{Suhr2020ExploringUG}. We test our models on four additional datasets and show that the weakly supervised models are generally better than the fully supervised ones in terms of cross-database generalization.
Overall, our findings show that weak supervision, in the form of question, answers and QDMRs (annotated or predicted) is nearly as effective as gold SQL when training text-to-SQL parsers. 

Our codebase and data are publicly available.\footnote{\url{https://github.com/tomerwolgithub/question-decomposition-to-sql}}

\section{Background}
\label{sec:background}


\paragraph{Weakly Supervised ML}
\label{sec:background_weak_supervision}

The performance of supervised ML models hinges on the quantity and quality of their training data. In practice, labeling large-scale datasets for new tasks is often cost-prohibitive. 
This problem is further exacerbated in semantic parsing tasks \cite{zettlemoyer05ccg}, as utterances need to be labeled with formal queries. 
\emph{Weak supervision} is a broad class of methods aimed at reducing the need to manually label large training sets \cite{hoffmann2011knowledge,Ratner2017SnorkelRT, Zhang2019LearningFI}.
An influential line of work has been dedicated to weakly supervised semantic parsing, using question-answer pairs, referred to as \emph{learning from denotations} \cite{clarke10world,
liang11dcs}. 
Past work has shown that non-experts are capable of annotating answers over knowledge graphs \cite{berant2013freebase} and tabular data \cite{pasupat2015compositional}. This approach could potentially be extended to databases by sampling subsets of its tables, such that question-answer examples can be manually annotated.
A key issue in learning text-to-SQL parsers from denotations is the vast search space of potential candidate queries. Therefore, past work has focused on constraining the search space, which limited applicability to simpler questions over single tables \cite{Wang2019LearningSP}. Here, we handle arbitrary SQL by using QDMR to constrain the search space.


\paragraph{Question Decomposition}
\label{sec:background_qdmr}
QDMR expresses the meaning of a question by breaking it down into simpler sub-questions. 
Given a question $x$, its QDMR $s$ is a sequence of reasoning steps $s^1, ..., s^{|s|}$ required to answer $x$. Each step $s^k$ is an intermediate question which represents a relational operation, such as projection or aggregation. Steps may contain phrases from $x$, tokens signifying a query operation (e.g., \textit{``for each''}) and references to previous steps. Operation tokens indicate the structure of a step, while its arguments are the references and question phrases.
A key advantage of QDMR is that it can be annotated by non-experts and at scale \cite{wolfson-etal-2020-break}. Moreover, unlike SQL, annotating QDMR requires zero domain expertise as it is derived entirely from the original question.

\section{Weakly Supervised SQL Synthesis}
\label{sec:data_generation}
Our input data contains examples of question $x_i$, database $D_i$, the answer $a_i$, and $s_i$, which is the QDMR of $x_i$. The QDMR is either annotated or predicted by a trained model $f$, such that $f(x_i) = s_i$. 
For each example, we attempt to synthesize a SQL query $\hat{Q}_i$, that matches the intent of $x_i$ and executes to its answer, $\hat{Q}_i(D_i)=a_i$.
The successfully synthesized examples $\langle x_i,\hat{Q}_i \rangle$ are then used to train a text-to-SQL model.

\subsection{Synthesizing SQL from QDMR}
Given QDMR $s_i$ and database $D_i$, we automatically map $s_i$ to SQL. Alg.~\ref{alg:grounding} describes the synthesis process, where candidate query $\hat{Q}_i$ is incrementally synthesized by iterating over the QDMR steps. Given step $s_i^k$, its phrases are automatically linked to columns and values in $D_i$. Then, relevant join paths are inferred between the columns. Last, each step is automatically mapped to SQL based on its QDMR operator and its arguments (see Table~\ref{tab:sql_mapper_example}). 

\begin{algorithm}[t]\setlength{\belowcaptionskip}{-8pt}
\scriptsize
\centering
\caption{SQL Synthesis}
\begin{algorithmic}[1]
\Procedure{SQLSynth}{$\mathbf{s}$: QDMR, $D$: database}
    \State $\textit{mapped}\gets []$
    \For{$s^k$ in $\mathbf{s} = \langle s^1, ..., s^n \rangle$}
        \State $\textit{cols} \gets\textsc{PhraseColumnLink}(D, s^k)$ \label{line:phrase_link}
        \State $\textit{refs} \gets\textsc{ReferencedSteps}(s^k)$ \label{line:ref_ext}
        \State $\textit{join}\gets []$
        \For{$s^j$ in $\textit{refs}$}
            \State $\textit{other\_cols}\gets \textit{mapped}[j].cols$
            \State $\textit{join}\gets \textit{join} + \textsc{JoinP}(D, \textit{cols}, \textit{other\_cols})$\label{line:join_path}
        \EndFor
        \State $op\gets\textsc{OpType}(s^k)$ \label{line:op_ext}
        \State $\hat{Q}\gets \textsc{MapSQL}(\textit{op, cols, join, refs, mapped})$ \label{line:mapper}
        \State $\textit{mapped}[k]\gets \langle s^k, cols, \hat{Q} \rangle$
    \EndFor
    \State \textbf{return} $\textit{mapped}[n].\hat{Q}$ 
\EndProcedure
\end{algorithmic}
\label{alg:grounding}
\end{algorithm}

\begin{table*}[t]\setlength{\belowcaptionskip}{-8pt}
\scriptsize
  \begin{tabular}{p{0.17\linewidth}p{0.22\linewidth}p{0.52\linewidth}}
    \toprule
    QDMR Step & Phrase-DB Linking & SQL \\ 
    \midrule
    1. ships &  1. \texttt{SELECT(ship.id})  & {\fontfamily{qcr}\selectfont SELECT ship.id FROM ship;} \vspace*{0.05cm}\\ 
    2. injuries &  2. \texttt{SELECT(death.injured)} & {\fontfamily{qcr}\selectfont SELECT death.injured FROM death;} \vspace*{0.05cm}\\ 
    3. number of \#2 for each \#1 &  3. \texttt{GROUP(count, \#2, \#1)} & {\fontfamily{qcr}\selectfont  SELECT COUNT(death.injured) FROM ship, death WHERE \underline{death.caused\_by\_ship\_id = ship.id} GROUP BY ship.id;}  \vspace*{0.05cm}\\ 
    4. \#1 where \#3 is highest & 4. \texttt{SUPER.(max, \#1, \#3)} & {\fontfamily{qcr}\selectfont SELECT ship.id FROM ship, death WHERE death.caused\_by\_ship\_id = ship.id GROUP BY ship.id ORDER BY COUNT(death.injured) DESC LIMIT 1;}\vspace*{0.05cm}\\ 
    5. the name of \#4  & 5. \texttt{PROJECT(ship.name, \#4)}  & {\fontfamily{qcr}\selectfont  SELECT ship.name FROM ship, death WHERE death.caused\_by\_ship\_id = ship.id AND ship.id IN (\#4);}  \\ 
  \bottomrule
\end{tabular}
\caption{Mapping the QDMR of the question \textit{``What is the ship name that caused most total injuries?''} to SQL.} 
\label{tab:sql_mapper_example}
\end{table*}

\subsubsection{Phrase DB Linking}
\label{sec:phrase_schema_linking}
As discussed in \S\ref{sec:background_qdmr}, a QDMR step may have a phrase from $x_i$ as its argument. When mapping QDMR to SQL these phrases are linked to corresponding columns or values in $D_i$. For example, in Table~\ref{tab:sql_mapper_example} the two phrases \textit{``ships''} and \textit{``injuries''} are linked to columns \texttt{ship.id} and \texttt{death.injured} respectively. We perform phrase-column linking automatically by ranking all columns in $D_i$ and returning the top one. The ranked list of columns is later used in \S\ref{sec:execution_guided_search} when searching for a correct assignment to all phrases in the QDMR. 
To compute phrase-column similarity, we tokenize both the phrase and column, then lemmatize their tokens using the WordNet lemmatizer.\footnote{\url{https://www.nltk.org/}} The similarity score is the average GloVe word embeddings similarity \cite{pennington2014glove} between the phrase and column tokens. All columns in $D_i$ are then ranked based on their word overlap and similarity with the phrase: (1) we return columns whose lemmatized tokens are identical to those in the phrase; (2) we return columns who share (non stop-word) tokens with the phrase, ordered by phrase-column similarity; (3) we return the remaining columns, ordered by similarity. 

We assume that literal values in $D_i$, such as strings or dates, appear verbatim in the database as they do in the question. Therefore, using string matching, we can identify the columns containing all literal values mentioned in $s_i$. If a literal value appears in multiple columns, they are all returned as potential links. 
This assumption may not always hold in practice due to DB-specific language, e.g., the phrase \textit{``women''} may correspond to the condition \texttt{gender = `F'}. Consequently, we measure the effect of DB-specific language in \S\ref{sec:experiments_data_generation}.


\subsubsection{Join Path Inference}
\label{sec:join_paths}
In order to synthesize SQL \cite{codd1970relational}, we infer join paths between the linked columns returned in \S\ref{sec:phrase_schema_linking}. Following past work \cite{GuoIRNet2019, Suhr2020ExploringUG}, we implement a heuristic returning the shortest join path connecting two sets of columns. To compute join paths, we convert $D_i$ into a graph where the nodes are its tables and edges exist for every foreign-key constraint connecting two tables.
The \textsc{JoinP} procedure in Alg.~\ref{alg:grounding} joins the tables of columns mentioned in step $s^k$ (\textit{cols}) with those mentioned in the previous steps which $s^k$ refers to (\textit{other\_cols}). If multiple shortest paths exist, it returns the first path which contains either $c_i \in \textit{cols}$ as its start node or $c_j \in \textit{other\_cols}$ as its end node. 
Step 3 of Table~\ref{tab:sql_mapper_example} underlines the join path between the \texttt{death} and \texttt{ship} tables.

\subsubsection{QDMR to SQL Mapper}
\label{sec:qdmr_sql_mapper}
The \textsc{MapSQL} procedure in Alg.~\ref{alg:grounding} maps QDMR steps into executable SQL. First, the QDMR operation of each step is inferred from its utterance template using the \textsc{OpType} procedure of \citet{wolfson-etal-2020-break}. Then, following the previous DB linking phase, the arguments of each step are either the linked columns and values or references to previous steps (second column of Table~\ref{tab:sql_mapper_example}). \textsc{MapSQL} uses the step operation type and arguments to automatically map $s^k$ to a SQL query $\hat{Q}^k$. Each operation has a unique mapping rule to SQL, as shown in Table~\ref{tab:qdmr_ops_to_sql}. SQL mapping is performed incrementally for each step. Then, when previous steps are referenced, the process can re-use parts of their previously mapped SQL (stored in the \textit{mapped} array). 
Furthermore, our mapping procedure is able to handle complex SQL that may involve nested queries (Fig.~\ref{fig:project_op_example}) and self-joins (Fig.~\ref{fig:self_join_example}).

\begin{table*}[t]\setlength{\belowcaptionskip}{-8pt}
\scriptsize
  \begin{tabular}{p{0.21\linewidth}p{0.73\linewidth}}
    \toprule
    QDMR Operation & SQL Mapping \\
    \midrule
    {\fontfamily{qcr}\selectfont SELECT(t.col)} & {\fontfamily{qcr}\selectfont SELECT t.col FROM t;} \\\midrule
    {\fontfamily{qcr}\selectfont FILTER(\#x, =, val)} & {\fontfamily{qcr}\selectfont SELECT \#x[SELECT] FROM \#x[FROM] WHERE \#x[WHERE] AND t.col = val;} \\ \midrule
    {\fontfamily{qcr}\selectfont PROJECT(t.col, \#x)} & {\fontfamily{qcr}\selectfont SELECT t.col FROM t, \#x[FROM] WHERE Join(t, \#x[FROM]) AND \#x[SELECT] IN (\#x);} \\\midrule
    {\fontfamily{qcr}\selectfont AGGREGATE(count, \#x)} & {\fontfamily{qcr}\selectfont SELECT COUNT(\#x[SELECT]) FROM \#x[FROM] WHERE \#x[WHERE];} \\\midrule
    {\fontfamily{qcr}\selectfont GROUP(avg, \#x, \#y)} & {\fontfamily{qcr}\selectfont SELECT AVG(\#x[SELECT]) FROM \#x[FROM], \#y[FROM] WHERE Join(\#x[FROM], \#y[FROM]) AND \#x[WHERE] AND \#y[WHERE] GROUP BY \#y[SELECT];} \\\midrule
    {\fontfamily{qcr}\selectfont SUPER.(max, k, \#x, \#y)} & {\fontfamily{qcr}\selectfont SELECT \#x[SELECT] FROM \#x[FROM], \#y[FROM] WHERE Join(\#x[FROM], \#y[FROM]) AND \#x[WHERE] AND \#y[WHERE] ORDER BY \#y[SELECT] DESC k;} \\
  \bottomrule
\end{tabular}
\caption{QDMR to SQL mapping rules for six operations. The full set of mapping rules, for all QDMR operations, is provided in the Appendix \ref{sec:appendix_qdmr_to_SQL}. \#x denotes a previously mapped SQL query while \#x[CLAUSE] denotes its relevant SQL clause. For example, \#x[FROM] returns all tables in the FROM clause of SQL query \#x. }
\label{tab:qdmr_ops_to_sql}
\end{table*}


\begin{figure}[t]\setlength{\belowcaptionskip}{-8pt}
\scriptsize
  \begin{tabular}{p{0.001\linewidth}p{0.87\linewidth}}
    \toprule
     $x$: & \textbf{{``What are the populations of states through which the Mississippi river runs?''}} \\
     \midrule
     $s$: & the Mississippi river; states \#1 runs through; the populations of \#2\\
     \midrule
     1. & {\fontfamily{qcr}\selectfont SELECT(river.river\_name = `Mississippi')}\\
     2. & {\fontfamily{qcr}\selectfont PROJECT(state.state\_name, \hl{\#1}) }\\
     3. & {\fontfamily{qcr}\selectfont PROJECT(state.population, \hl{\#2})}\\
     \midrule
     1. & {\fontfamily{qcr}\selectfont SELECT river.river\_name FROM river WHERE river.river\_name = `Mississippi';} \\
     2. & {\fontfamily{qcr}\selectfont SELECT state.state\_name FROM state, river WHERE river.traverse = state.state\_name AND river.river\_name IN (\hl{\#1});} \\
     3. & {\fontfamily{qcr}\selectfont SELECT state.population FROM state, river WHERE river.traverse = state.state\_name AND state.state\_name IN (\hl{\#2});} \\ 
  \bottomrule
\end{tabular}
\caption{Previously mapped QDMR steps (with operations and arguments) used as nested SQL queries.}
\label{fig:project_op_example}
\end{figure}


\begin{figure}[t]\setlength{\belowcaptionskip}{-8pt}
\scriptsize
  \begin{tabular}{p{0.001\linewidth}p{0.87\linewidth}}
    \toprule
     $x$: & \textbf{``What papers were written by both H. V. Jagadish and also Yunyao Li?''} \\
     \midrule
     $s$: & papers; \#1 by H. V. Jagadish; \#2 by Yunyao Li\\
     \midrule
     1. & {\fontfamily{qcr}\selectfont SELECT(publication.title)}\\
     2. & {\fontfamily{qcr}\selectfont FILTER(\#1, \hl{author.name = `H. V. Jagadish'})}\\
     3. & {\fontfamily{qcr}\selectfont FILTER(\#2, \hl{author.name = `Yunyao Li'})}\\
     \midrule
     1. & {\fontfamily{qcr}\selectfont SELECT publication.title FROM author, publication;} \\ 
     2. & {\fontfamily{qcr}\selectfont SELECT publication.title FROM author, publication, writes WHERE publication.pid= writes.pid AND writes.aid = author.aid AND author.name = `H. V. Jagadish'; } \\ 
     3. & {\fontfamily{qcr}\selectfont SELECT publication.title FROM author, publication, writes WHERE publication.pid = writes.pid AND writes.aid = author.aid AND author.name = `Yunyao Li' AND publication.title IN (\hl{\#2});} \\ 
  \bottomrule
\end{tabular}
\caption{Handling Self-joins in QDMR mapping. The two \texttt{FILTER} steps have conflicting assignments to the same column and are identified as a self-join. This is resolved by using a nested query in the SQL of step 3. }
\label{fig:self_join_example}
\end{figure}

\subsection{Execution-guided SQL Candidate Search}
\label{sec:execution_guided_search}
At this point we have $\hat{Q}_i$, which is a potential SQL candidate. However, this candidate may be incorrect due to a wrong phrase-column linking or due to its original QDMR structure. To mitigate these issues, we search for accurate SQL candidates using the answer supervision.

Following phrase DB linking (\S\ref{sec:phrase_schema_linking}), each phrase is assigned its top ranked column in $D_i$. However, this assignment may potentially be wrong. In step 1 of Fig.~\ref{fig:architecture} the phrase \textit{``authors''} is incorrectly linked to \texttt{author.aid} instead of \texttt{author.name}. Given our weak supervision, we do not have access to the gold phrase-column linking and rely instead on the gold answer $a_i$. Namely, we iterate over phrase-column assignments and synthesize their corresponding SQL. Once an assignment whose SQL executes to $a_i$ has been found, we return it as our result. We iterate over assignments that cover only the top-k ranked columns for each phrase, shown to work very well in practice (\S\ref{sec:synthesis_annotated_qdmr}).


Failing to find a correct candidate SQL may be due to QDMR structure rather than phrase-column linking. As $s_i$ is derived entirely from the question it may fail to capture database-specific language. E.g., in the question \textit{``How many students enrolled during the semester?''} the necessary aggregate operation may change depending on the database structure. If $D_i$ has the column \texttt{course.num\_enrolled}, the query should \emph{sum} its entries for all courses in the semester. Conversely, if $D_i$ has the column \texttt{course.student\_id}, the corresponding query would need to \emph{count} the number of enrolled students.
We account for these structural mismatches by implementing three additional search heuristics which modify the structure of a candidate $\hat{Q}_i$. If the candidate executes to the correct result following modification, it is returned by the search process. 
These modifications are described in detail in Appendix \ref{sec:appendix_sql_candidate_search}. Namely, they include the addition of a \texttt{DISTINCT} clause, converting QDMR \texttt{FILTER} steps into \texttt{SUPERLATIVE} and switching between the \texttt{COUNT} and \texttt{SUM} operations.

\section{Experiments}
\label{sec:experiments}
Our experiments target two main research questions. 
First, given access to weak supervision of question-answer pairs and QDMRs, we wish to measure the percentage of SQL queries that can be automatically synthesized. Therefore, in \S\ref{sec:experiments_data_generation} we measure SQL synthesis \emph{coverage} using 9,313 examples taken from five benchmark datasets. Second, in \S\ref{sec:experiments_nl_to_sql} we use the synthesized SQL to train text-to-SQL models and compare their performance to those trained on gold SQL annotations.


\subsection{Setting}
\label{sec:experiments_setting}
\paragraph{Datasets} 
We evaluate both the SQL synthesis coverage and text-to-SQL accuracy using five text-to-SQL datasets (see Table~\ref{tab:grounding}). The first four datasets contain questions over a single database: \academic{} \cite{Li2014NaLIRAI} has questions over the Microsoft Academic Search database; \geo{} \cite{zelle96geoquery} concerns US geography; \imdb{} and \yelp{} \cite{Yaghmazadeh2017SQLizerQS} contain complex questions on a film and restaurants database, respectively. The \spider{} dataset \cite{Yu2018SpiderAL} measures \emph{domain generalization} between databases, and therefore contains questions over 160 different databases.
For QDMR data we use the \breakdata{} dataset \cite{wolfson-etal-2020-break}. The only exception is 259 questions of \imdb{} and \yelp{}, outside of \breakdata{}, which we (authors) annotate with corresponding QDMRs and release with our code.
See Appendix \ref{sec:appendix_dataset_license} for license.

\paragraph{Training} 
\label{sec:experiments_setting_training}
We fine-tune the \tfive{} sequence-to-sequence model \cite{Raffel2020ExploringTL} for both text-to-SQL and QDMR parsing ($\S\ref{sec:experiments_data_generation_predictions}$). Namely, for each task we fine-tune the pre-trained model on its specific data. 
For text-to-SQL, we fine-tune on mapping utterances $x_i;cols(D_i)$ to SQL, where sequence $cols(D_i)$ is a serialization of all columns in database $D_i$, in an arbitrary order. In QDMR parsing, input questions are mapped to output QDMR strings.
We use the T5 implementation by HuggingFace \cite{wolf-etal-2020-transformers} and train using the Adam optimizer \cite{kingma2014adam}. Following fine-tuning on the dev sets, we adjust the batch size to 128 and the learning rate to $1e$-4 (after experimenting with $1e$-5, $1e$-4 and $1e$-3). All models were trained on an NVIDIA GeForce RTX 3090 GPU.

\subsection{SQL Synthesis Coverage}
\label{sec:experiments_data_generation}
Our first challenge is to measure our ability to synthesize accurate SQL.  
To evaluate SQL synthesis we define its \emph{coverage} to be the percentage of examples where it successfully produces SQL $\hat{Q}$ which executes to the correct answer.
To ensure our procedure is domain independent, we test it on five different datasets, spanning 164 databases (Table~\ref{tab:grounding}). 

\paragraph{Annotated QDMRs}
\label{sec:synthesis_annotated_qdmr}
The upper rows of Table~\ref{tab:grounding} list the SQL synthesis coverage when using manually annotated QDMRs from \breakdata{}. Overall, we evaluate on 9,313 QDMR annotated examples, reaching a coverage of 77.8\%. Synthesis coverage for single-DB datasets tends to be slightly higher than that of \spider{}, which we attribute to its larger size and diversity. 
To further ensure the quality of synthesized SQL, we manually validate a random subset of 100 queries out of the 7,249 that were synthesized. Our analysis reveals 95\% of the queries to be correct interpretations of their original question. 
In addition, we evaluate synthesis coverage on a subset of 8,887 examples whose SQL denotations are not the empty set. As SQL synthesis relies on answer supervision, discarding examples with empty denotations eliminates the false positives of spurious SQL which incidentally execute to an empty set. Overall, coverage on examples with non-empty denotations is nearly identical, at 77.6\% (see Appendix~\ref{sec:appendix_sql_synthesis_nonempty}). 
We also perform an error analysis on a random set of 100 failed examples, presented in Table~\ref{tab:grounding_errors}. SQL synthesis failures are mostly due to QDMR annotation errors or implicit database-specific conditions. E.g., in \geo{} the phrase \textit{``major river''} should implicitly be mapped to the condition \texttt{river.length > 750}. As our SQL synthesis is domain-general, it does not memorize any domain-specific jargon or rules. 

\begin{table}[t]\setlength{\belowcaptionskip}{-5pt}
\centering
\scriptsize
  \begin{tabular}{lcccc}
    \toprule
    Dataset & DB \# & Examples & Synthesized & Coverage \% \\
    \midrule
    \academic{} & 1 & 195 & 155 & 79.5 \\
    \geo{} & 1 & 877 & 736 & 83.9 \\
    \imdb{} & 1 & 131 & 116 & 88.5 \\
    \yelp{} & 1  & 128 & 100 & 78.1 \\
    \spider{} dev & 20 & 1,027 & 793 & 77.2 \\
    \spider{} train & 140 & 6,955 & 5,349 & 76.9 \\
    \bf Total: & \bf 164 & \bf 9,313 & \bf 7,249 & \bf 77.8 \\
    \midrule
    \spider{} pred. & 20 & 1,027 & 797 & 77.6 \\
  \bottomrule
\end{tabular}
\caption{SQL synthesis coverage scores across datasets.}
  \label{tab:grounding}
\end{table}


\begin{table}[t]\setlength{\belowcaptionskip}{-8pt}
\scriptsize
\centering
  \begin{tabular}{p{0.16\linewidth}p{0.625\linewidth}c}
    \toprule
    Error & Description & \% \\
    \midrule
    Nonstandard QDMR & The annotated QDMR contains a step utterance that does not follow any of the pre-specified operation templates in \citet{wolfson-etal-2020-break} & 42  \\\midrule
    DB-specific language & Phrase entails an implicit condition, e.g., \textit{``female workers''} $\rightarrow$ {\fontfamily{qcr}\selectfont emp.gender = `F'} & 23  \\\midrule
    Phrase-column link. & The correct phrase-column assignment falls outside of the top-k candidates (\S\ref{sec:execution_guided_search}) & 13  \\\midrule
    Gold SQL & An error due to an incorrectly labeled gold SQL & 6  \\
  \bottomrule
\end{tabular}
\caption{SQL synthesis error analysis.}
  \label{tab:grounding_errors}
\end{table}

\paragraph{Predicted QDMRs}
\label{sec:experiments_data_generation_predictions}
While QDMR annotation can be crowdsourced to non-experts \cite{wolfson-etal-2020-break}, moving to a new domain may incur annotating new in-domain examples.
Our first step to address this issue is to evaluate the coverage of SQL synthesis on predicted QDMRs, for out-of-domain questions. As question domains in \spider{} dev differ from those in its training set, it serves as our initial testbed. We further explore this setting in \S\ref{sec:experiments_nl_to_sql_predicted_qdmr}.
Our QDMR parser (\S\ref{sec:experiments_setting_training}) is fine-tuned on \breakdata{} for 10 epochs and we select the model with highest exact string match (EM) on \breakdata{} dev. 
Evaluating on the hidden test set reveals our model scores 42.3 normalized EM,\footnote{The metric is a strict lower bound on performance.} setting the state-of-the-art on \breakdata{}.\footnote{\url{https://leaderboard.allenai.org/break}}
The predicted QDMRs, are then used in SQL synthesis together with examples $\langle x_i, a_i, D_i\rangle$. In Table~\ref{tab:grounding}, the last row shows that coverage on \spider{} dev is nearly identical to that of manually annotated QDMRs (77.6\% to 77.2\%). 


\subsection{Training Text-to-SQL Models}
\label{sec:experiments_nl_to_sql}
Next, we compare text-to-SQL models trained on our synthesized data to training on expert annotated SQL.
Given examples $\langle x_i, D_i \rangle$ we test the following settings:
(1) A \emph{fully supervised} training set, that uses gold SQL annotations $\{\langle x_i, Q_i, D_i \rangle\}_{i=1}^{n}$.
(2) A \emph{weakly supervised} training set, where given answer $a_i$ and QDMR $s_i$, we synthesize queries $\hat{Q}_i$. As SQL synthesis coverage is less than 100\%, the process returns a subset of $m<n$ examples $\{\langle x_i, \hat{Q}_i, D_i \rangle\}_{i=1}^{m}$ on which the model is trained.\footnote{In practice, we do not train directly on $\hat{Q}_i$ but on $s_i$ following its phrase-column linking. This representation is then automatically mapped to SQL to evaluate its execution.}

\subsubsection{Training Data} 
We train models on two text-to-SQL datasets: \spider{} \cite{Yu2018SpiderAL} and \geo{} \cite{zelle96geoquery}.
As our weakly supervised training sets, we use the synthesized examples $\langle x_i, \hat{Q}_i, D_i \rangle$, described in \S\ref{sec:experiments_data_generation}, (using annotated QDMRs). We successfully synthesized 5,349 training examples for \spider{} and 547 examples for \geo{} train.

\subsubsection{Models and Evaluation} 
\paragraph{Models} 
We fine-tune \tfive{} for text-to-SQL, using the hyperparameters from \S\ref{sec:experiments_setting_training}. We choose T5 as it is agnostic to the structure of its input sequences. Namely, it has been shown to perform competitively on different text-to-SQL datasets, regardless of their SQL conventions \cite{Shaw2021CompositionalGA, Herzig2021UnlockingCG}. This property is particularly desirable in our cross-database evaluation (\S\ref{sec:experiments_nl_to_sql_annotated_qdmr}), where we test on multiple datasets.

We train and evaluate the following models:
\begin{itemize}[leftmargin=*,topsep=0pt,itemsep=0pt,parsep=0pt]
  \item \textbf{\tfivegold{}} trained on $\{\langle x_i, Q_i, D_i \rangle\}_{i=1}^{n}$, using \emph{gold SQL}, annotated by experts
  \item \textbf{\tfiveqdmr{}} trained on $\{\langle x_i, \hat{Q}_i, D_i \rangle\}_{i=1}^{m}$ with $\hat{Q}_i$ that were synthesized using weak supervision  
  \item \textbf{\tfivegoldpartial{}} trained on $\{\langle x_i, Q_i, D_i \rangle\}_{i=1}^{m}$, using gold SQL. This models helps us measure the degree to which the smaller size of the synthesized training data and its different query structure (compared to gold SQL) affects performance
\end{itemize}


\paragraph{Evaluation Metric} 
Due to our SQL being automatically synthesized, its syntax is often different from that of the gold SQL (see Appendix~\ref{sec:appendix_nl2sql_additional_results_qualitative}). As a result, the ESM metric of \citet{Yu2018SpiderAL} does not fit our evaluation setup. Instead, we follow \citet{Suhr2020ExploringUG} and evaluate text-to-SQL models using the \emph{execution accuracy} of output queries.   
We define execution accuracy as the percentage of output queries which, when executed on the database, result in the same set of tuples (rows) as $a_i$.

\begin{table}[t]\setlength{\belowcaptionskip}{-5pt}
\centering
\scriptsize
  \begin{tabular}{lcccc}
    \toprule
    Model & Supervision & Training set & Exec. \% \\
    \midrule
    \tfivegold{} & $\langle x_i, Q_i, D_i \rangle$ & 7,000 & 68.0 $\pm$ 0.3 \\
    \tfivegoldpartial{} & $\langle x_i, Q_i, D_i \rangle$ & 5,349 & 66.4 $\pm$ 0.8 \\
    \tfiveqdmr{} & $\langle x_i, a_i, s_i, D_i \rangle$ & 5,349 & 65.8 $\pm$ 0.3 \\
    \tfivepred{} & $\langle x_i, a_i,D_i \rangle$* & 5,075 & 62.9 $\pm$ 0.8 \\
  \bottomrule
\end{tabular}
\caption{\spider{} trained model results on the dev set. *Supervision for \tfivepred{} also includes 700 annotated QDMRs of \spider{} train questions.}
  \label{tab:t5_spider}
\end{table}

\begin{table}[t]\setlength{\belowcaptionskip}{-8pt}
\centering
\scriptsize
  \begin{tabular}{lccccccc}
    \toprule
    Model & \academic{} & \geo{} & \imdb{} & \yelp{}\\
    \midrule
    \tfivegold{} & 8.2 $\pm$ 1.3 & 33.6 $\pm$ 2.5 & 19.8 $\pm$ 3.6 & 22.7 $\pm$ 1.2 \\
    \tfivegoldpartial{} & 4.9 $\pm$ 1.5 & 32.4 $\pm$ 1.3 & 20.9 $\pm$ 0.8 & 20.7 $\pm$ 1.4 \\
    \tfiveqdmr{} & 10.7 $\pm$ 0.7 & 40.4 $\pm$ 1.8 & 27.1 $\pm$ 3.6 & 16.2 $\pm$ 4.7  \\
    \tfivepred{} & 8.2 $\pm$ 0.4 & 39.7 $\pm$ 1.4 & 23.6 $\pm$ 5.5 & 16.7 $\pm$ 3.7 \\
  \bottomrule
\end{tabular}
\caption{\spider{} trained models zero-shot performance on cross-database (XSP) examples.}
  \label{tab:t5_spider_xsp}
\end{table}

\subsubsection{Training on Annotated QDMRs} 
\label{sec:experiments_nl_to_sql_annotated_qdmr}
We begin by comparing the models trained using annotated QDMRs to those trained on gold SQL. Meanwhile, the discussion of \tfivepred{}, trained using predicted QDMRs, is left for \S\ref{sec:experiments_nl_to_sql_predicted_qdmr}.
The results in Tables~\ref{tab:t5_spider}-\ref{tab:t5_geo} list the average accuracy and standard deviation of three model instances, trained using separate random seeds. 

\paragraph{\spider{} \& XSP Evaluation}
Tables~\ref{tab:t5_spider}-\ref{tab:t5_spider_xsp} list the results of the \spider{} trained models. 
All models were trained for 150 epochs and evaluated on the dev set of 1,034 examples. When comparing \tfiveqdmr{} to the model trained on gold SQL, it achieves 96.8\% of its performance (65.8 to 68.0). 
The \tfivegoldpartial{} model, trained on the same 5,349 examples as \tfiveqdmr{}, performs roughly on par, scoring +0.6 points (66.4 to 65.8).

As \spider{} is used to train cross-database models, we further evaluate our models performance on \emph{cross-database semantic parsing} (XSP) \cite{Suhr2020ExploringUG}. In Table~\ref{tab:t5_spider_xsp} we test on four additional text-to-SQL datasets (sizes in parenthesis): \academic{} (183), \geo{} (877), \imdb{} (113) and \yelp{} (66). For \academic{}, \imdb{} and \yelp{} we removed examples whose execution result in an empty set. Otherwise, the significant percentage of such examples would result in false positives of predictions which incidentally execute to an empty set. In practice, evaluation on the full datasets remains mostly unchanged and is provided in  Appendix~\ref{sec:appendix_nl2sql_additional_results}.
Similarly to \citet{Suhr2020ExploringUG}, the results in Table~\ref{tab:t5_spider_xsp} show that \spider{} trained models struggle to generalize to XSP examples. However, \tfiveqdmr{} performance is generally better on XSP examples, which further indicates that QDMR and answer supervision is effective, compared to gold SQL. Example predictions are shown in Appendix~\ref{sec:appendix_nl2sql_additional_results_qualitative}.

\paragraph{\geo{}} 
Table~\ref{tab:t5_geo} lists the execution accuracy of models trained on \geo{}. Models were trained for 300 epochs, fine-tuned on the dev set and then evaluated on the 280 test examples. We note that \tfiveqdmr{} achieves 90.7\% of the performance of \tfivegold{} (74.5 to 82.1). The larger performance gap, compared to \spider{} models, may be partly to due to the dataset size. As \geo{} has 547 training examples, fewer synthesized SQL to train \tfiveqdmr{} on (454) may have had a greater effect on its accuracy.

\begin{table}[t]\setlength{\belowcaptionskip}{-8pt}
\centering
\scriptsize
  \begin{tabular}{lcccc}
    \toprule
    Model & Supervision & Train. set & Exec. \% \\
    \midrule
    \tfivegold{} & $\langle x_i, Q_i, D_i \rangle$ & 547 & 82.1 $\pm$ 1.9 \\
    \tfivegoldpartial{} & $\langle x_i, Q_i, D_i \rangle$ & 454 & 79.4 $\pm$ 0.4  \\
    \tfiveqdmr{} & $\langle x_i, a_i, s_i, D_i \rangle$ & 454 & 74.5 $\pm$ 0.2  \\
    \tfivepred{} & $\langle x_i, a_i, D_i \rangle$ & 432 & 70.4 $\pm$ 0.2  \\
  \bottomrule
\end{tabular}
\caption{\geo{} trained models results on the test set. Supervision for \tfivepred{} does not include any in-domain annotated QDMRs.}
  \label{tab:t5_geo}
\end{table}

\subsubsection{Training on Predicted QDMRs} 
\label{sec:experiments_nl_to_sql_predicted_qdmr}
We extend our approach by replacing the annotated QDMRs with the predictions of a trained QDMR parser (a \tfive{} model, see \S\ref{sec:experiments_setting_training}).
In this setting, we now have two sets of questions: (1) questions used to train the QDMR parser; (2) questions used to synthesize NL-SQL data.
We want these two sets to be as separate as possible, so that training the QDMR parser would not require new \emph{in-domain} annotations. Namely, the parser must generalize to questions in the NL-SQL domains while being trained on as few of these questions as possible.

\paragraph{\spider{}}
Unfortunately, \spider{} questions make up a large portion of the \breakdata{} training set, used to train the QDMR parser. We therefore experiment with two alternatives to minimize the in-domain QDMR annotations of NL-SQL questions. First, is to train the parser using few-shot QDMR annotations for \spider{}. Second, is to split \spider{} to questions used as the NL-SQL data, while the rest are used to train the QDMR parser.

In Table~\ref{tab:t5_spider}, \tfivepred{} is trained on 5,075 queries, synthesized using predicted QDMRs (and answer supervision) for \spider{} train questions. 
The predictions were generated by a QDMR parser trained on a subset of \breakdata{}, excluding all \spider{} questions save 700 (10\% of \spider{} train). Keeping few in-domain examples minimizes additional QDMR annotation while preserving the predictions quality. Training on the predicted QDMRs, instead of the annotated ones, resulted in accuracy being down 2.9 points (65.8 to 62.9) while the model achieves 92.5\% of \tfivegold{} performance on \spider{} dev. On XSP examples, \tfivepred{} is competitive with \tfiveqdmr{} (Table~\ref{tab:t5_spider_xsp}).

In Table~\ref{tab:t5_spider_predicted}, we experiment with training \tfivepred{} without in-domain QDMR annotations. We avoid any overlap between the questions and domains used to train the QDMR parser and those used for SQL synthesis. We randomly sample 30-40 databases from \spider{} and use their corresponding questions exclusively as our NL-SQL data. For training the QDMR parser, we use \breakdata{} while discarding the sampled questions. 
We experiment with 3 random samples of \spider{} train, numbering 1,348, 2,028 and 2,076 examples, with synthesized training data of 1,129, 1,440 and 1,552 examples respectively. Results in Table~\ref{tab:t5_spider_predicted} show that, on average, \tfivepred{} achieves 95.5\% of the performance of \tfivegold{}. This indicates that even without any in-domain QDMR annotations, data induced from answer supervision and out-of-domain QDMRs is effective in training text-to-SQL models, compared to gold SQL.

\paragraph{\geo{}}
For predicted QDMRs on \geo{}, we train the QDMR parser on \breakdata{} while discarding all of its 547 questions. Therefore, the parser was trained without any in-domain QDMR annotations for \geo{}. SQL synthesis using the predicted QDMRs resulted in 432 queries. In Table~\ref{tab:t5_geo}, \tfivepred{} reaches 85.7\% of \tfivegold{} performance while being trained using question-answer supervision and no in-domain QDMR annotations.

\begin{table}[t]\setlength{\belowcaptionskip}{-8pt}
\centering
\scriptsize
  \begin{tabular}{lcccc}
    \toprule
    Model & Supervision & Train. set & DB \# & Exec. \% \\
    \midrule
    \tfivegold{} & $\langle x_i, Q_i, D_i \rangle$ & 1,348 & 30 & 48.4  \\
    \tfivegoldpartial{} & $\langle x_i, Q_i, D_i \rangle$ & 1,129 & 30 & 47.4  \\
    \tfivepred{} & $\langle x_i, a_i, D_i \rangle$ & 1,129 & 30 & 46.2  \\
    \midrule
    \tfivegold{} & $\langle x_i, Q_i, D_i \rangle$ & 2,028 & 40 & 54.7  \\
    \tfivegoldpartial{} & $\langle x_i, Q_i, D_i \rangle$ & 1,440 & 40 & 51.3  \\
    \tfivepred{} & $\langle x_i, a_i, D_i \rangle$ & 1,440 & 40 & 52.1  \\
    \midrule
    \tfivegold{} & $\langle x_i, Q_i, D_i \rangle$ & 2,076 & 40 & 56.2  \\
    \tfivegoldpartial{} & $\langle x_i, Q_i, D_i \rangle$ & 1,552 & 40 & 53.7  \\
    \tfivepred{} & $\langle x_i, a_i, D_i \rangle$ & 1,552 & 40 & 53.8  \\
  \bottomrule
\end{tabular}
\caption{\spider{} models results on the dev set. \tfivepred{} is trained without using any QDMR annotations for training set questions. We train separate models on the three randomly sampled training sets.}
  \label{tab:t5_spider_predicted}
\end{table}

\section{Limitations} 
\label{sec:limitations}
Our approach uses question decompositions and answers as supervision for text-to-SQL parsing. As annotating SQL requires expertise, our solution serves as a potentially cheaper alternative. Past work has shown that non-experts can provide the answers for questions on knowledge graphs \cite{berant2013freebase} and tables \cite{pasupat2015compositional}. However, manually annotating question-answer pairs on large-scale databases may present new challenges which we leave for future work.

During SQL synthesis we assume that literal values (strings or dates) appear verbatim in the database as they do in the question. We observe that, for multiple datasets, this assumption generally holds true (\S\ref{sec:experiments_data_generation}). Still, for questions with domain-specific jargon \cite{lee-etal-2021-kaggledbqa} our approach might require an initial step of named-entity-recognition.
Failure to map a QDMR to SQL may be due to a mismatch between a QDMR and its corresponding SQL structure (\S\ref{sec:execution_guided_search}). We account for such mismatches by using heuristics to modify the structure of a candidate query (Appendix \ref{sec:appendix_sql_candidate_search}). A complementary approach could train a model, mapping QDMR to SQL, to account for cases where our heuristic rules fail. Nevertheless, our SQL synthesis covers a diverse set of databases and query patterns, as shown in our experiments.

\section{Related Work}
\label{sec:related_work}

For a thorough review of NL interfaces to databases see \citet{Affolter2019ACS, Kim2020NaturalLT}. 
Research on parsing text-to-SQL gained significant traction in recent years with the introduction of large supervised datasets for training models and evaluating their performance \cite{zhong2017seq2sql, Yu2018SpiderAL}. Recent approaches relied on specialized architectures combined with pre-trained language models \cite{GuoIRNet2019, Wang2020RATSQLRS, lin-etal-2020-bridging, yu2020Grappa, Deng2021StructureGroundedPF, scholak-etal-2021-picard}. As our solution synthesizes NL-SQL pairs (using weak supervision) it can be used to train supervised text-to-SQL models. 

Also related is the use of intermediate meaning representations (MRs) in mapping text-to-SQL. In past work MRs were either annotated by experts \cite{Yaghmazadeh2017SQLizerQS, Kapanipathi2020QuestionAO}, or were directly induced from such annotations as a way to simplify the target MR \cite{dong-lapata-2018-coarse,GuoIRNet2019,Herzig2021UnlockingCG}. Instead, QDMR representations are expressed as NL utterances and can therefore be annotated by non-experts.
Similarly to us, \citet{Saparina2021SPARQLingDQ} map QDMR to SPARQL. However, our SQL synthesis does not rely on the annotated linking of question phrases to DB elements \cite{Lei2020ReexaminingTR}. In addition, we train models without gold QDMR annotations and test our models on four datasets in addition to \spider{}.

\section{Conclusions}
This work presents a weakly supervised approach for generating NL-SQL training data, using answer and QDMR supervision.
We implemented an automatic SQL synthesis procedure, capable of generating effective training data for dozens of target databases.
Experiments on multiple text-to-SQL benchmarks demonstrate the efficacy of our synthesized training data. Namely, our weakly-supervised models achieve 91\%-97\% of the performance of fully supervised models trained on annotated SQL. Further constraining our models supervision to few or zero in-domain QDMRs still reaches 86\%-93\% of the fully supervised models performance.
Overall, we provide an effective solution to train text-to-SQL parsers while requiring zero SQL annotations.


\section*{Acknowledgements}
We would like to thank Mor Geva, Ori Yoran and Jonathan Herzig for their insightful comments. 
This research was partially supported by The Israel Science Foundation (grant 978/17), and the Yandex Initiative for Machine Learning and the European Research Council (ERC) under the European Union Horizons 2020 research and innovation programme (grants DELPHI 802800 and ProDIS 804302).
This work was completed in partial fulfillment of the PhD of Tomer Wolfson.

\bibliography{anthology,custom}
\bibliographystyle{acl_natbib}

\appendix

\section{QDMR to SQL Mapping Rules}
\label{sec:appendix_qdmr_to_SQL}

Table~\ref{tab:qdmr_ops_to_sql_full} lists all of the QDMR operations along with their mapping rules to SQL. For a thorough description of QDMR semantics please refer to \citet{wolfson-etal-2020-break}.

\begin{table*}[t]
\tiny
  \begin{tabular}{p{0.24\linewidth}p{0.7\linewidth}}
    \toprule
    QDMR Operation & SQL Mapping \\
    \midrule
    {\fontfamily{qcr}\selectfont SELECT(t.col)} & {\fontfamily{qcr}\selectfont SELECT t.col FROM t;} \\\midrule
    {\fontfamily{qcr}\selectfont SELECT(val)} & {\fontfamily{qcr}\selectfont SELECT t.col FROM t WHERE t.col = val;} \\\midrule
    {\fontfamily{qcr}\selectfont FILTER(\#x, =, val)} & {\fontfamily{qcr}\selectfont SELECT \#x[SELECT] FROM \#x[FROM] WHERE \#x[WHERE] AND t.col = val;} \\ \midrule
    {\fontfamily{qcr}\selectfont PROJECT(t.col, \#x)} & {\fontfamily{qcr}\selectfont SELECT t.col FROM t, \#x[FROM] WHERE Join(t, \#x[FROM]) AND \#x[SELECT] IN (\#x);} \\\midrule
    {\fontfamily{qcr}\selectfont AGGREGATE(count, \#x)} & {\fontfamily{qcr}\selectfont SELECT COUNT(\#x[SELECT]) FROM \#x[FROM] WHERE \#x[WHERE];} \\\midrule
    {\fontfamily{qcr}\selectfont GROUP(avg, \#x, \#y)} & {\fontfamily{qcr}\selectfont SELECT AVG(\#x[SELECT]) FROM \#x[FROM], \#y[FROM] WHERE Join(\#x[FROM], \#y[FROM]) AND \#x[WHERE] AND \#y[WHERE] GROUP BY \#y[SELECT];} \\\midrule
    {\fontfamily{qcr}\selectfont SUPERLATIVE(max, k, \#x, \#y)} & {\fontfamily{qcr}\selectfont SELECT \#x[SELECT] FROM \#x[FROM], \#y[FROM] WHERE Join(\#x[FROM], \#y[FROM]) AND \#x[WHERE] AND \#y[WHERE] ORDER BY \#y[SELECT] DESC k;} \\\midrule
    {\fontfamily{qcr}\selectfont COMPARATIVE(\#x, \#y, >, val)} & {\fontfamily{qcr}\selectfont SELECT \#x[SELECT] FROM \#x[FROM], \#y[FROM] WHERE Join(\#x[FROM], \#y[FROM]) AND \#x[WHERE] AND \#y[WHERE] AND \#y[SELECT] > val;} \\\midrule
    {\fontfamily{qcr}\selectfont UNION(\#x, \#y)} & {\fontfamily{qcr}\selectfont SELECT \#x[SELECT] FROM \#x[FROM], \#y[FROM] WHERE Join(\#x[FROM], \#y[FROM]) AND (\#x[WHERE] OR \#y[WHERE]);} \\\midrule
    {\fontfamily{qcr}\selectfont UNION\_COLUMN(\#x, \#y)} & {\fontfamily{qcr}\selectfont SELECT \#x[SELECT], \#y[SELECT] FROM \#x[FROM], \#y[FROM] WHERE Join(\#x[FROM], \#y[FROM]) AND \#x[WHERE] AND \#y[WHERE];} \\\midrule
    {\fontfamily{qcr}\selectfont INTERSECT(t.col, \#x, \#y)} & {\fontfamily{qcr}\selectfont SELECT t.col FROM t, \#x[FROM], \#y[FROM] WHERE Join(t, \#x[FROM], \#y[FROM]) AND \#x[WHERE] AND t.col IN ( SELECT t.col FROM t, \#x[FROM], \#y[FROM] WHERE Join(t, \#x[FROM], \#y[FROM]) AND \#y[WHERE] );} \\\midrule
    {\fontfamily{qcr}\selectfont SORT(\#x, \#y, asc)} & {\fontfamily{qcr}\selectfont SELECT \#x[SELECT] FROM \#x[FROM], \#y[FROM] WHERE Join(\#x[FROM], \#y[FROM]) AND \#x[WHERE] ORDER BY \#y[SELECT] ASC;} \\\midrule
    {\fontfamily{qcr}\selectfont DISCARD(\#x, \#y)} & {\fontfamily{qcr}\selectfont SELECT \#x[SELECT] FROM \#x[FROM] WHERE \#x[WHERE] AND \#x[SELECT] NOT IN ( \#y );} \\\midrule
    {\fontfamily{qcr}\selectfont ARITHMETIC(+, \#x, \#y)} & {\fontfamily{qcr}\selectfont ( \#x ) + ( \#y );} \\
  \bottomrule
\end{tabular}
\caption{QDMR to SQL mapping rules for all QDMR operations. \#x denotes a previously mapped SQL query while \#x[CLAUSE] denotes its relevant SQL clause. For example, \#x[FROM] returns all tables in the FROM clause of SQL query \#x. Join, denotes the inferred join paths between sets of tables (see \S\ref{sec:join_paths}). Note that \texttt{AGGREGATE} and \texttt{GROUP} steps may use the operations: min, max, count, sum and avg. \texttt{SUPERLATIVE} steps may use  min, max operations and \texttt{COMPARATIVE} steps use the operations: $>$, $<$, $=$, $\neq$, $\geq$, $\leq$. Last, \texttt{SORT} steps sort in either ascending (asc) or descending (desc) order and \texttt{ARITHMETIC} steps use one of the following: $+$, $-$, $\times$, $\div$.}
\label{tab:qdmr_ops_to_sql_full}
\end{table*}

\section{SQL Candidate Search Heuristics}
\label{sec:appendix_sql_candidate_search}

We further describe the execution-guided search process for candidate SQL queries, that was introduced in \S\ref{sec:execution_guided_search}. Given the search space of candidate queries, we use four heuristics to find candidates $\hat{Q}_i$ which execute to the correct answer, $a_i$.

\vspace{2mm}
\noindent\textbf{1. Phrase linking search:} 
We avoid iterating over each phrase-column assignment by ordering them according to their phrase-column ranking, as described in \S\ref{sec:phrase_schema_linking}. The query $\hat{Q}_i^{(1)}$ is induced from the top ranked assignment, where each phrase in $s_i$ is assigned its top ranked column. If $\hat{Q}_i^{(1)}(D_i) \neq a_i$ we continue the candidate search using heuristics 2-4 (described below). Assuming that the additional search heuristics failed to find a candidate $\hat{Q}_i^{(1)'}$ such that $\hat{Q}_i^{(1)'}(D_i) = a_i$, we return to the phrase linking component and resume the process using the candidate SQL induced from the following assignment $\hat{Q}_i^{(2)}$, and so forth. In practice, we limit the number of assignments and review only those covering the top-k most similar columns for each phrase in $s_i$, where $k=20$. Our error analysis (Table~\ref{tab:grounding_errors}) reveals that only a small fraction of failures are due to limiting $k$. Step 2 in Fig.~\ref{fig:architecture} represents the iterative process, where $\hat{Q}_i^{(1)}$ executes to an incorrect result while the following candidate $\hat{Q}_i^{(2)}$ correctly links the phrase \textit{``authors''} to column \texttt{author.name} and executes to $a_i$, thereby ending the search.

\vspace{2mm}
\noindent\textbf{2. Distinct modification:} 
Given a candidate SQL $\hat{Q}_i$ such that $\hat{Q}_i(D_i) \neq a$, we add \texttt{DISTINCT} to its \texttt{SELECT} clause. In Table~\ref{tab:synthesis_search_heuristics} the SQL executes to the correct result, following its modification.

\vspace{2mm}
\noindent\textbf{3. Superlative modification:} 
This heuristic automatically corrects semantic mismatches between annotated QDMR structures and the underlying database. Concretely, steps in $s_i$ that represent \texttt{PROJECT} and \texttt{FILTER} operations may entail an implicit \texttt{ARGMAX/ARGMIN} operation. For example for the question \textit{``What is the size of the largest state in the USA?''} in the third row of Table~\ref{tab:synthesis_search_heuristics}. Step (3) of the question's annotated QDMR is the \texttt{PROJECT} operation, ``state with the largest \#2''. While conforming to the \texttt{PROJECT} operation template, the step entails an \texttt{ARGMAX} operation. Using the NLTK part-of-speech tagger, we automatically identify any superlative tokens in the \texttt{PROJECT} and \texttt{FILTER} steps of $s_i$. These steps are then replaced with the appropriate \texttt{SUPERLATIVE} type steps. In Table~\ref{tab:synthesis_search_heuristics}, the original step (3) is modified to the step ``\#1 where \#2 is highest''.

\vspace{2mm}
\noindent\textbf{4. Aggregate modification:} 
This heuristics replaces instances of \texttt{COUNT} in QDMR steps with \texttt{SUM} operations, and vice-versa. In Table~\ref{tab:synthesis_search_heuristics}, the question \textit{``Find the total student enrollment for different affiliation type schools.''}, is incorrectly mapped to a candidate query involving a \texttt{COUNT} operation on \texttt{university.enrollment}. By modifying the aggregate operation to \texttt{SUM}, the new $\hat{Q}_i$ correctly executes to $a_i$ and is therefore returned as the output.

\begin{table*}[t]
\scriptsize
  \begin{tabular}{p{0.07\linewidth}p{0.19\linewidth}p{0.31\linewidth}p{0.31\linewidth}}
    \toprule
    Heuristic & Question & Candidate SQL/QDMR & Modified Candidate SQL/QDMR \\ 
    \midrule
    Phrase linking search & What are the distinct majors that students with treasurer votes are studying? & {\fontfamily{qcr}\selectfont SELECT DISTINCT student.major FROM student, voting\_record WHERE student.stuid = \textbf{voting\_record.stuid}} & {\fontfamily{qcr}\selectfont SELECT DISTINCT student.major FROM student, voting\_record WHERE student.stuid = \textbf{voting\_record.treasurer\_vote}} \\ 
    \midrule
    Distinct modification & Find the number of different product types. & {\fontfamily{qcr}\selectfont SELECT products.product\_type\_code FROM products}  & {\fontfamily{qcr}\selectfont SELECT \textbf{DISTINCT} products.product\_type\_code FROM products} \\ 
    \midrule
    Superlative modification & What is the size of the largest state in the USA? & {\fontfamily{qcr}\selectfont (1) states in the usa; (2) size of \#1; \textbf{(3) state with the largest \#2}; (4) size of \#3}  & {\fontfamily{qcr}\selectfont (1) states in the usa; (2) size of \#1; \textbf{(3) \#1 where \#2 is highest}; (4) the size of \#3} \\ 
    \midrule
    Aggregate modification & Find the total student enrollment for different affiliation type schools. & {\fontfamily{qcr}\selectfont SELECT university.affiliation, \textbf{COUNT}(university.enrollment) FROM university GROUP BY university.affiliation} & {\fontfamily{qcr}\selectfont SELECT university.affiliation, \textbf{SUM}(university.enrollment) FROM university GROUP BY university.affiliation} \\ 

  \bottomrule
\end{tabular}
\caption{Examples of the four execution-guided search heuristics used during SQL synthesis.}
  \label{tab:synthesis_search_heuristics}
\end{table*}

\section{Data License}
\label{sec:appendix_dataset_license}
We list the license (when publicly available) and the release details of the datasets used in our paper.

The \breakdata{} dataset \cite{wolfson-etal-2020-break} is under the MIT License.
\spider{} \cite{Yu2018SpiderAL} is under the CC BY-SA 4.0 License. \geo{} \cite{zelle96geoquery} is available under the GNU General Public License 2.0.

The text-to-SQL versions of \geo{} and \academic{} \cite{Li2014NaLIRAI} were made publicly available by \citet{finegan-dollak2018improving} in: \url{https://github.com/jkkummerfeld/text2sql-data/}. 

The \imdb{} and \yelp{} datasets were publicly released by \citet{Yaghmazadeh2017SQLizerQS} in: \url{goo.gl/DbUBMM}.

\section{SQL Synthesis Coverage}
\label{sec:appendix_sql_synthesis_nonempty}
We provide additional results of SQL synthesis coverage. Table~\ref{tab:grounding_nonempty} lists the coverage results, per dataset, when discarding all examples whose SQL executes to an empty set. Out of the 9,313 original examples, 8,887 examples have non-empty denotations. Coverage scores per dataset remain generally the same as they do when evaluating on all examples. These results further indicate the effectiveness of the SQL synthesis procedure. Namely, this ensures the synthesis results in Table~\ref{tab:grounding} are faithful, despite the potential noise introduced by SQL with empty denotations.

\begin{table}[t]
\centering
\scriptsize
  \begin{tabular}{lcccc}
    \toprule
    Dataset & DB \# & Examples & Synthesized & Coverage \% \\
    \midrule
    \academic{} & 1 & 183 & 148 & 80.9 \\
    \geo{} & 1 & 846 & 707 & 83.6 \\
    \imdb{} & 1 & 113 & 101 & 89.4 \\
    \yelp{} & 1 & 66 & 54 & 81.8 \\
    \spider{} dev & 20 & 978 & 745 & 76.2 \\
    \spider{} train & 140 & 6,701 & 5,137 & 76.7 \\
    \bf Total: & \bf 164 & \bf 8,887 & \bf 6,892 & \bf 77.6 \\
    \midrule
    \spider{} pred. & 20 & 978 & 750 & 76.7 \\
  \bottomrule
\end{tabular}
\caption{SQL synthesis coverage scores for SQL queries with non-empty denotations. We report the coverage only for non-empty examples to minimize the effect of potentially spurious SQL being synthesized.}
  \label{tab:grounding_nonempty}
\end{table}

\section{NL to SQL Models Results}
\label{sec:appendix_nl2sql_additional_results}
\subsection{Evaluation on the Full XSP Datasets}
We provide additional results of the models trained on \spider{}. Namely, we evaluate on all examples of the \academic{}, \imdb{} and \yelp{} datasets, including examples whose denotations are empty.
Table~\ref{tab:t5_spider_xsp_empty_denotations} lists the results of all the models trained on the original training set of \spider{}.
In Table~\ref{tab:t5_spider_predicted_xsp} we provide the XSP results of the models trained on the random subsets of \spider{} train, used in \S\ref{sec:experiments_nl_to_sql_predicted_qdmr}. Similar to our previous experiments, \tfivepred{} is generally better than \tfivegold{} in terms of its cross-database generalization.

\subsection{Qualitative Results}
\label{sec:appendix_nl2sql_additional_results_qualitative}
Table~\ref{tab:nl_to_sql_qualitative} includes some example predictions of our \spider{} trained models from Tables~\ref{tab:t5_spider}-\ref{tab:t5_spider_xsp}. For each example we describe its question and target (gold) SQL annotation, followed by each model's result.

\begin{table*}[t]
\centering
\scriptsize
  \begin{tabular}{lcccccccc}
    \toprule
    Model & Supervision & Training set & \spider{} dev. & \academic{} & \geo{} & \imdb{} & \yelp{} \\
    \midrule
    \tfivegold{} & $\langle x_i, Q_i, D_i \rangle$ & 7,000 & 68.0 $\pm$ 0.3 & 7.9 $\pm$ 1.3 & 33.6 $\pm$ 2.5 & 19.1 $\pm$ 2.9 & 25.3 $\pm$ 1.7 \\
    \tfivegoldpartial{} & $\langle x_i, Q_i, D_i \rangle$ & 5,349 & 66.4 $\pm$ 0.8 & 4.9 $\pm$ 1.7 & 32.4 $\pm$ 1.3 & 21.1 $\pm$ 0.7 & 26.1 $\pm$ 1.0 \\
    \tfiveqdmr{} & $\langle x_i, a_i, s_i, D_i \rangle$ & 5,349 & 65.8 $\pm$ 0.3 & 11.2 $\pm$ 1.0 & 40.4 $\pm$ 1.8 & 30.3 $\pm$ 3.1 & 25.8 $\pm$ 5.1  \\
    \tfivepred{} & $\langle x_i, a_i,D_i \rangle$ & 5,075 & 62.9 $\pm$ 0.8 & 8.4 $\pm$ 0.9 & 39.7 $\pm$ 1.4 & 27.0 $\pm$ 5.1 & 28.2 $\pm$ 2.9  \\
  \bottomrule
\end{tabular}
\caption{Model execution accuracy on \spider{} and its performance on cross-database (XSP) examples. Evaluation on \academic{}, \imdb{} and \yelp{} is on the \emph{full datasets}, including examples with empty denotations.}
  \label{tab:t5_spider_xsp_empty_denotations}
\end{table*}


\begin{table*}[t]
\centering
\scriptsize
  \begin{tabular}{lcccccccc}
    \toprule
    Model & Supervision & Train. set & DB \# & \spider{} dev. & \academic{} & \geo{} & \imdb{} & \yelp{} \\
    \midrule
    \tfivegold{} & $\langle x_i, Q_i, D_i \rangle$ & 1,348 & 30 & 48.4 & 2.1 & 29.6 & 9.9 & 22.6 \\
    \tfivegoldpartial{} & $\langle x_i, Q_i, D_i \rangle$ & 1,129 & 30 & 47.4 & 2.6 & 26.9 & 14.5 & 16.9 \\
    \tfivepred{} & $\langle x_i, a_i, D_i \rangle$ & 1,129 & 30 & 46.2  & 8.4 & 29.0 & 16.0 & 16.9 \\
    \midrule
    \tfivegold{} & $\langle x_i, Q_i, D_i \rangle$ & 2,028 & 40 & 54.7  & 6.3 & 28.3 & 18.3 & 21.0 \\
    \tfivegoldpartial{} & $\langle x_i, Q_i, D_i \rangle$ & 1,440 & 40 & 51.3  & 3.7 & 21.2 & 12.2 & 19.4 \\
    \tfivepred{} & $\langle x_i, a_i, D_i \rangle$ & 1,440 & 40 & 52.1  & 6.8 & 27.4 & 12.2 & 18.5 \\
    \midrule
    \tfivegold{} & $\langle x_i, Q_i, D_i \rangle$ & 2,076 & 40 & 56.2  & 3.2 & 25.5 & 13.0 & 24.5 \\
    \tfivegoldpartial{} & $\langle x_i, Q_i, D_i \rangle$ & 1,552 & 40 & 53.7  & 2.3 & 17.8 & 10.2 & 22.8 \\
    \tfivepred{} & $\langle x_i, a_i, D_i \rangle$ & 1,552 & 40 & 53.8  & 6.1 & 32.3 & 19.8 & 21.8 \\

  \bottomrule
\end{tabular}
\caption{Model results on \spider{} dev when trained on predicted QDMRs versus gold SQL. We train separate models on each of the three randomly sampled training sets. Results include the performance on XSP examples where the evaluation on \academic{}, \imdb{} and \yelp{} is on the \emph{full datasets}, including examples with empty denotations.}
  \label{tab:t5_spider_predicted_xsp}
\end{table*}

\begin{table*}[t]
\centering
\scriptsize
  \begin{tabular}{p{0.1\linewidth}p{0.7\linewidth}p{0.01\linewidth}}
    \toprule
     \bf Question: & Return me the total citations of papers in the VLDB conference in 2005. & \\
     \bf Target SQL: & {\fontfamily{qcr}\selectfont select sum ( publication\_0.citation\_num ) from conference as conference\_0, publication as publication\_0 where conference\_0.name = "VLDB" and publication\_0.year = 2005 and conference\_0.cid = publication\_0.cid;}\vspace*{0.1cm} & \\
     \bf \tfivegold{}: & {\fontfamily{qcr}\selectfont select sum(t1.citation\_num) from publication as t1 join conference as t2 on t1.cid = t2.cid where t2.name = "VLDB" and t1.year < 2005;}\vspace*{0.1cm} & \cmark \\
     \bf \tfiveqdmr{}: & {\fontfamily{qcr}\selectfont SELECT SUM(cite.cited) FROM publication, cite, conference WHERE conference.cid = publication.cid AND publication.pid = cite.cited AND conference.cid IN ( SELECT conference.cid FROM conference WHERE conference.name = 'VLDB' ) AND publication.year < 2005;}\vspace*{0.1cm} & \xmark \\
     \bf \tfivepred{}: & {\fontfamily{qcr}\selectfont SELECT SUM(cite.cited) FROM cite, publication, conference WHERE publication.pid = cite.cited AND conference.cid = publication.cid AND conference.cid IN ( SELECT conference.cid FROM publication, conference WHERE conference.cid = publication.cid AND conference.name = 'VLDB' AND publication.year < 2005 ); }\vspace*{0.1cm} & \xmark \\
     \midrule
     \bf Question: & Return me the papers written by H. V. Jagadish and Yunyao Li after 2005. & \\
     \bf Target SQL: & {\fontfamily{qcr}\selectfont select publication\_0.title from author as author\_0, author as author\_1, publication as publication\_0, writes as writes\_0, writes as writes\_1 where author\_0.name = "Yunyao Li" and author\_1.name = "H. V. Jagadish" and publication\_0.year  >  2005 and author\_0.aid = writes\_0.aid and author\_1.aid = writes\_1.aid and publication\_0.pid = writes\_0.pid and publication\_0.pid = writes\_1.pid;}\vspace*{0.1cm} & \\
     \bf \tfivegold{}: & {\fontfamily{qcr}\selectfont select t3.title from author as t1 join writes as t2 on t1.aid = t2.aid join publication as t3 on t2.citation\_num = t3.pid where t1.name = "H. V. Jagadish" intersect select t3.title from author as t1 join writes as t2 on t1.aid = t2.aid join journal as t3 on t2.journal = t3.jid where t3.name = "Yunyao Li";}\vspace*{0.1cm} & \xmark \\
     \bf \tfiveqdmr{}: & {\fontfamily{qcr}\selectfont SELECT publication.abstract FROM author, publication, writes WHERE publication.pid = writes.pid AND writes.aid = author.aid AND author.name = 'h.v.jagadish' AND publication.year > 2005}\vspace*{0.1cm} & \xmark \\
     \bf \tfivepred{}: & {\fontfamily{qcr}\selectfont SELECT journal.homepage FROM journal, author, publication, writes WHERE journal.jid = publication.jid AND publication.pid = writes.pid AND writes.aid = author.aid AND author.name = 'h.v.jagadish' AND publication.year > 2005 }\vspace*{0.1cm} & \xmark \\
     \midrule
     \bf Question: & What is the shortest river? & \\
     \bf Target SQL: & {\fontfamily{qcr}\selectfont SELECT river.river\_name FROM river WHERE river.length = (SELECT min (river.length) FROM river);}\vspace*{0.1cm} & \\
     \bf \tfivegold{}: & {\fontfamily{qcr}\selectfont select length from river order by length limit 1;}\vspace*{0.1cm} & \xmark \\
     \bf \tfiveqdmr{}: & {\fontfamily{qcr}\selectfont SELECT river.river\_name FROM river WHERE river.river\_name IN ( SELECT river.river\_name FROM river ) ORDER BY river.length ASC LIMIT 1;}\vspace*{0.1cm} & \cmark \\
     \bf \tfivepred{}: & {\fontfamily{qcr}\selectfont SELECT river.river\_name FROM river WHERE river.river\_name IN ( SELECT river.river\_name FROM river ) ORDER BY river.length ASC LIMIT 1; }\vspace*{0.1cm} & \cmark \\
     \midrule
     \bf Question: & Find the actor with most number of films. & \\
     \bf Target SQL: & {\fontfamily{qcr}\selectfont select actor\_0.name from actor as actor\_0, cast as cast\_0, movie as movie\_0 where actor\_0.aid = cast\_0.aid and cast\_0.msid = movie\_0.mid order by count(distinct(movie\_0.title)) desc limit 1;}\vspace*{0.1cm} & \\
     \bf \tfivegold{}: & {\fontfamily{qcr}\selectfont select t1.name from actor as t1 join cast as t2 on t1.aid = t2.id group by t1.aid order by count(*) desc limit 1;}\vspace*{0.1cm} & \xmark \\
     \bf \tfiveqdmr{}: & {\fontfamily{qcr}\selectfont SELECT actor.name FROM actor, movie, cast WHERE actor.aid = cast.aid AND cast.msid = movie.mid  GROUP BY actor.name ORDER BY COUNT(movie.mid) DESC LIMIT 1;}\vspace*{0.1cm} & \cmark \\
     \bf \tfivepred{}: & {\fontfamily{qcr}\selectfont SELECT actor.name FROM actor, cast, movie WHERE actor.aid = cast.aid AND cast.msid = movie.mid  GROUP BY actor.name ORDER BY COUNT(movie.mid) DESC LIMIT 1; }\vspace*{0.1cm} & \cmark \\
     \midrule
     \bf Question: & Which business has the most number of checkins? & \\
     \bf Target SQL: & {\fontfamily{qcr}\selectfont select business\_0.name from business as business\_0, checkin as checkin\_0 where business\_0.business\_id = checkin\_0.business\_id group by business\_0.name order by sum(checkin\_0.count) desc limit 1;}\vspace*{0.1cm} & \\
     \bf \tfivegold{}: & {\fontfamily{qcr}\selectfont select t1.name from business as t1 join checkin as t2 on t1.business\_id = t2.business\_id group by t2.business\_id order by count(*) desc limit 1;}\vspace*{0.1cm} & \xmark \\
     \bf \tfiveqdmr{}: & {\fontfamily{qcr}\selectfont SELECT business.name FROM checkin, business WHERE business.business\_id = checkin.business\_id  GROUP BY business.name ORDER BY COUNT(checkin.cid) DESC LIMIT 1;}\vspace*{0.1cm} & \xmark \\
     \bf \tfivepred{}: & {\fontfamily{qcr}\selectfont  SELECT business.name FROM checkin, business WHERE business.business\_id = checkin.business\_id  GROUP BY business.name ORDER BY COUNT(checkin.cid) DESC LIMIT 1;}\vspace*{0.1cm} & \xmark \\
    
  \bottomrule
\end{tabular}
\caption{Example predictions of the \spider{} trained models from Tables~\ref{tab:t5_spider}-\ref{tab:t5_spider_xsp}. We denote correct and incorrect predictions by \cmark and \xmark.}
\label{tab:nl_to_sql_qualitative}
\end{table*}

\end{document}